\pdfoutput=1

\PassOptionsToPackage{table}{xcolor}

\documentclass
{article}

\usepackage{times}
\usepackage{latexsym}

\usepackage[T1]{fontenc}

\usepackage[utf8]{inputenc}

\usepackage{microtype}

\usepackage{inconsolata}

\usepackage{graphicx}

\usepackage{subfigure}
\usepackage{booktabs} 

\usepackage{hyperref}
\usepackage{url}

\usepackage{amssymb}
\usepackage{mathtools}
\usepackage{amsthm}

\usepackage[preprint]{acl}

\usepackage[table]{xcolor}
\usepackage{listings}
\usepackage{adjustbox}

\usepackage
[
capitalize,
noabbrev
]
{cleveref}

\usepackage[
textsize=small,
disable
]{todonotes}

\usepackage{xurl}

\usetikzlibrary {shapes.geometric}

\usepackage{amsmath,amsfonts,bm}

\def\eqref#1{equation~\ref{#1}}

\def\1{\bm{1}}

\def\vw{{\bm{w}}}
\def\vx{{\bm{x}}}

\def\mW{{\bm{W}}}

\DeclareMathAlphabet{\mathsfit}{\encodingdefault}{\sfdefault}{m}{sl}
\SetMathAlphabet{\mathsfit}{bold}{\encodingdefault}{\sfdefault}{bx}{n}

\theoremstyle{plain}

\theoremstyle{definition}

\theoremstyle{remark}

\def\win{\vw_{\text{in}}}
\def\wgate{\vw_{\text{gate}}}
\def\wout{\vw_{\text{out}}}
\def\xin{x_{\text{in}}}
\def\xgate{x_{\text{gate}}}
\def\xpost{x_{\text{post}}}
\def\shortpar#1{\textbf{#1.}}

\long\def\devour#1{}

\newcounter{notecounter}
\newcommand{\enotesoff}{\long\gdef\enote##1##2{}}
\newcommand{\enoteson}{\long\gdef\enote##1##2{{
\stepcounter{notecounter}
{\large\bf
\hspace{0cm}\arabic{notecounter} $<<<$ ##1: ##2
$>>>$\hspace{1cm}}}}}

\enoteson
\enotesoff

\def\neuroscope{GLUScope}

\title{
	\neuroscope:\\
	A Resource for Analyzing GLU Neurons in Transformer Language Models
}

\author{Sebastian Gerstner, Leonor Veloso, Hinrich Schütze \\
  LMU Munich
  and Munich Center for Machine Learning (MCML)\\
  \texttt{sgerstner at cis dot lmu dot country}
  }

\begin{document}

\maketitle

\begin{abstract}
	We present \textbf{\neuroscope},
	an open-source resource for analyzing neurons
	in Transformer-based language models (LMs),
	intended for interpretability researchers.
	We focus on the now-standard \textit{GLU-based activation functions} such as SwiGLU,
	used in virtually all recent open-weights LMs.
	This introduces a new challenge:
	understanding positive activations is not enough.
	Instead, both the \textit{gate} and the \textit{in} activation of a neuron can be positive or negative,
	leading to four different possible sign
	combinations (\textit{activation quadrants})
	that in some cases have quite
	different functionalities
	(e.g. promoting vs. suppressing a specific set of tokens).
	Accordingly, for any neuron, our tool shows text examples for each of the four sign combinations,
	and indicates how often each combination occurs.
	Our resource can lead to novel insights:
	in particular,
	thanks to analyzing all four activation quadrants,
	it can surface interpretable behavior that other tools would miss.
	We also release the code used to create the resource, thus enabling researchers to build on it by adding more models, text corpora, or neuron pages.\footnote{
	Some sections of this paper overlap with our paper \citet{Gerstner2026Weakeningneurons}.
	\citet{Gerstner2026Weakeningneurons} reports on research findings, whereas the present paper presents the resource that enabled some of these findings.
	The overlap affects
	(a) \cref{sec:glu,sec:related-neuron} (which just provide background information),
	and (b) \cref{sec:examples} (where we describe how GLUScope enables some of the findings of \citealp{Gerstner2026Weakeningneurons}).
}
\end{abstract}

\section{Introduction}

Transformer-based 
\citep{Vaswani2017Attentionisall} large language models (LLMs)
have shown impressive performance,
but there is still much that is unclear about how they achieve it.
This had led to large-scale interest and effort
in understanding their inner workings,
an endeavor initially called
\textit{NLP interpretability} and later
\textit{mechanistic interpretability} \citep{Elhage2021mathematicalframeworktransformer,saphra-wiegreffe-2024-mechanistic}.
As a part of this endeavor,
several works have analyzed individual neurons,\footnote{
	We use "neuron" to refer to a hidden dimension inside the MLP layer.
}
especially via text examples that strongly activate them
\citep{geva-etal-2021-transformer, Nanda2022neuroscope, voita-etal-2024-neurons,Gurnee2024}.

We want to support these interpretability researchers,
especially those working on neuron interpretation,
by publishing a new resource for neuron analysis.
There are several such resources already (see below, \cref{sec:related}),
but there is a crucial gap to fill:
modern language models (e.g., OLMo \citealp{groeneveld-etal-2024-olmo,Olmo3} and Llama \citealp{Touvron2023,Llama3.2})
often use \textit{GLU-based} (a.k.a. \textit{gated}) \textit{activation functions}
such as SwiGLU or GEGLU \citep{Shazeer2020}.
These activation functions introduce a new challenge:
understanding positive activations is not enough.
Instead, both the \textit{gate} and the \textit{in}
activation of a neuron can be both positive and negative,
leading to four different possible sign combinations, which we call \textit{activation quadrants} (\cref{ss:quadrants}).
Each of these quadrants will have different patterns of appearance.
For this reason, GLU-based neurons can have much more complex behavior than non-GLU ones.
Our resource, \neuroscope{}, is the first to fully take this complexity into account:
For each neuron, we record not only the strongest activations overall,
but the strongest activations within each activation quadrant.
This surfaces interpretable behavior that other tools \citep{Nanda2022neuroscope,Choi2024Scalingautomaticneuron} would miss (\cref{sec:neuron-example}).

Our main contributions are the following:
\begin{enumerate}
	\item We release
	a range of datasets (for four different models) containing summary information on the activations of each neuron (\cref{sec:activation-data}).
	\item We release a website, \neuroscope{},
	for visualizing this data for a few selected neurons (\cref{sec:neuroscope}).
	\item We release the code we used to create the above, thus enabling other researchers to
	create similar activation datasets for other models and corpora, paving the way for future discoveries (\cref{sec:code}).
	\item We showcase our resource by two usage examples (\cref{sec:examples})
	that lead to new insights into neuron behavior,
	including ones that could not have been achieved with other resources.
\end{enumerate}

\section{GLU variants\footnote{
This section closely follows section 3 of \citet{Gerstner2026Weakeningneurons}.
}}
\label{sec:glu}

We work on \textit{GLU variants}
(a.k.a. \textit{gated activation functions})
like SwiGLU or GEGLU \citep{Shazeer2020}.
These activation functions are used widely; e.g., 
OLMo \citep{groeneveld-etal-2024-olmo,Olmo3}
and Llama
\citep{Touvron2023,Llama3.1,Llama3.2}
use SwiGLU,
and Gemma
\citep{Gemma2024,Gemma3}
uses GEGLU.

Traditional activation functions like ReLU
require just two weight matrices, $\mW_\text{in}$ and $\mW_\text{out}$.
If the pre-MLP residual stream state is $\vx$, the MLP outputs
\[\text{MLP}(\vx) := \mW_\text{out} \text{ReLU}(\mW_\text{in} \vx),\]
where ReLU is applied element-wise to each neuron
(it takes a single scalar as argument).
Other traditional activation functions are
$\text{Swish}(x):= x / (1+\exp(-x))$ \citep{Ramachandran2017}
and $\text{GELU}(x) := x \Phi(x)$\footnote{
	$\Phi$ is the cumulative distibution function (cdf) of a standard normal distribution.
}
\citep{Hendrycks2016}.
Both of these can be seen as smooth approximations of ReLU.

In contrast to these traditional functions, a \emph{gated activation function}
like SwiGLU
requires an additional weight matrix on the input side, called $\mW_\text{gate}$:
\begin{equation}
	\text{MLP}(\vx) = \mW_\text{out} \left(\sigma(\mW_{\text{gate}} \vx) \odot (\mW_{\text{in}} \vx)\right),
	\label{eq:mlp}
\end{equation}
where $\odot$ denotes element-wise multiplication (a.k.a. Hadamard product),
and $\sigma$ denotes the underlying activation function, which is Swish for SwiGLU and GELU for GEGLU.\footnote{
	In other works,
	$\mW_\text{in}$ and $\mW_\text{out}$
	are also called
	$\mW_\text{up}$ and $\mW_\text{down}$, respectively.
}

We find it more intuitive to separately consider each neuron:
The neuron adds the vector
\begin{equation}
	\sigma(\langle\wgate,\vx\rangle) \cdot \langle\win,\vx\rangle \cdot \wout
	\label{eq:neuron}
\end{equation}
to the residual stream.
Here $\wgate,\win$ are one of the $d_\text{MLP}$ \textit{rows} of $\mW_\text{gate}, \mW_\text{in}$,
and $\wout$ is one of the $d_\text{MLP}$ \textit{columns} of $\mW_\text{out}$.\footnote{
	This is assuming the right-to-left notation of our \cref{eq:mlp}.
	With left-to-right notation rows and columns are switched.
}
These weight vectors, as well as $\vx$, all have dimensionality $d_\text{model}$.
In this framework, SwiGLU can be described as a function of two scalars:
\[\mbox{SwiGLU}(\xgate, \xin) := \mbox{Swish}(\xgate) \cdot \xin,\]
where $\xgate := \langle\wgate,\vx\rangle$ and $\xin := \langle\win,\vx\rangle$.
We use $\xpost$ for the neuron activation, i.e.,
$\xpost = \text{SwiGLU}(\xgate, \xin)$.

\subsection{Activation quadrants}
\label{ss:quadrants}

Unlike ReLU, gated activation functions can output arbitrary positive or negative values.
For example, if $\xgate>0$ and $\xin\ll0$, then $\text{SwiGLU}(\xgate,\xin)\ll0$.

In fact the situation is even more complex than just positive or negative activations:
For a given neuron the $\xgate$ and $\xin$ values can take four possible sign combinations, which we call \textbf{activation quadrants}.
We denote these quadrants as \texttt{gate+\_in+} (for $\xgate>0$ and $\xin>0$),
\texttt{gate+\_in-}, etc.

Each of these quadrants will have different patterns of appearance.
For this reason, GLU-based neurons can have much more complex behavior than non-GLU ones.
Our resource, \neuroscope{}, is the first to take this complexity into account:
For each neuron, we record not only the strongest activations overall,
but the strongest activations within each activation quadrant.

\section{Related work}
\label{sec:related}
\subsection{Neuron analysis research\footnote{
		This section overlaps with section 2 of \citet{Gerstner2026Weakeningneurons}.
}}
\label{sec:related-neuron}
Much research has attempted to understand individual
neurons in transformer-based LMs.
Examples include
\citet{Miller2023,Dai2022Knowledgeneuronspretrained,Niu2024,Kong2026Negativepreactivations}.

Broadly speaking,
neuron analysis works can be divided into three distinct perspectives:
input-centered, output-centered, and input-output.
\textit{Input-centered} analyses are possibly the most frequent ones:
they attempt to understand what causes a neuron to activate.
Typically they run the model on a dataset and record which examples (most strongly) activate a given neuron.
Examples include \citet{geva-etal-2021-transformer, Nanda2022neuroscope, voita-etal-2024-neurons,Gurnee2024}.
\textit{Output-centered} analyses attempt to understand the effect of a neuron,
for example by an analysis of their output weight.
Examples include \citet{Gurnee2024,Stolfo2024}.
Finally, \textit{input-output} analyses attempt to understand the relationship between input and output of a neuron.
\citet{geva-etal-2021-transformer,gur-arieh-etal-2025-enhancing} do this at the semantic level,
and
\citet{Gerstner2026Weakeningneurons}
do it at the basic mathematical level by computing cosine similarities of input and output weights,
an idea that was also briefly mentioned in \citet{Elhage2021mathematicalframeworktransformer,Gurnee2024}.

The focus on individual neurons has been criticized.
\citet{Millidge2022}
find interpretable directions that do not correspond to individual neurons.
\citet{Elhage2022Toymodelssuperposition} argue that interpretable features 
are non-orthogonal directions in model space
and can be superposed.
This corresponds to sparse linear combinations of neurons in MLP space.
This has inspired a series of work on sparse autoencoders (SAEs), starting with \citet{Sharkey2022}.

The focus on SAEs has  been criticized:
recent studies indicate that they do not always outperform baselines \citep{Kantamneni2025,Leask2025,Mueller2025,Wu2025,Arora2026Languagemodelcircuits}.
A middle ground is possible:
\citet{Gurnee2023}
argue that interpretable features correspond to sparse combinations of neurons;
this includes 1-sparse combinations, i.e., individual neurons.
Accordingly, there is recent work on new classes of interpretable neurons, e.g., \citet{Ali2025Detectingpruningprominent, Zhao2025Findingculturesensitive}.

\subsection{Neuron analysis resources}

\paragraph{Neuroscope}
Our work is largely inspired by Neuroscope \cite{Nanda2022neuroscope}.
This is a website visualizing the top activations of each neuron from a range of models.
However it provides little additional information and is not kept up-to-date with newer models.
In particular, no GLU models are shown.

\paragraph{Transluce's neuron descriptions}
\citet{Choi2024Scalingautomaticneuron} publish top activations (and LLM-generated explanations) for a selection of neurons in Llama-3.1-8B-instruct \citep{Llama3.1}, a GLU-based model. (Their neuron pages are at \url{https://neurons.transluce.org}.)
They observed that, in GLU-based activation functions, negative activations can be just as relevant as positive ones.
Therefore they have separate pages for the positive and negative activations of each neuron.
We go one step further and consider four different activation quadrants for each neuron.

\paragraph{Other resources}
NeuroX \citep{Dalvi2019NeuroX}
is primarily a code base:
users can train their own (comparatively small) models,
run them on a (small) dataset
and then visualize neuron activations.

LM Debugger \citep{Geva2022}
has a different focus:
given a model prediction,
it shows the neurons that are most prominently activated on it,
together with an output-based analysis.
The user can adjust neuron activations to "debug" the particular model prediction.

Transformer Debugger \citep{Mossing2024}
is a more general set of interpretability tools for small language models
(focusing on GPT2-small \cite{Radford2019});
one of these tools is \textit{Neuron viewer}.
For a given neuron,
it shows text examples for activations in various quantile ranges
(e.g., from the top $0.1\%$ or bottom $50\%$ activations),
and tentative explanations generated by GPT-4.
Additionally, the tool shows the top tokens of input and output weights,
as well as related neurons (apparently based on weight similarity).

Neuronpedia \citep{Lin2023} is a large-scale platform for exploring latent features within language models,
for example those produced by sparse autoencoders (SAE, \citealp{Sharkey2022}).
It is however not directly designed for exploring neurons.
In contrast, we are specifically focusing on neurons.
For reasons outlined in \cref{sec:related-neuron},
we believe this is still a valuable endeavor.

None of the above tools addresses the problem we want to solve:
an interface to visualize neuron activations of GLU-based models,
that considers all four activation quadrants.

\subsection{Related code libraries}
Our work relies on two libraries:
First, TransformerLens \cite{Nanda2022transformerlens}
enables easy access to transformer hidden states.
We use the fork by \citet{Gerstner2026Weakeningneurons}, which implements an additional weight preprocessing step as described in their appendix D.

The second library is CircuitsVis \cite{Cooney2023}.
It provides functions for appealing visualizations,
such as coloring tokens according to their activation.
Our own visualization code is adapted from CircuitsVis.

Another visualization library,
that we did not use but deserves mention,
is Ecco \citep{alammar-2021-ecco}.

\section{Released artifacts}
\label{sec:artifacts}

\begin{figure}
	\includegraphics
	[width=\linewidth]
	{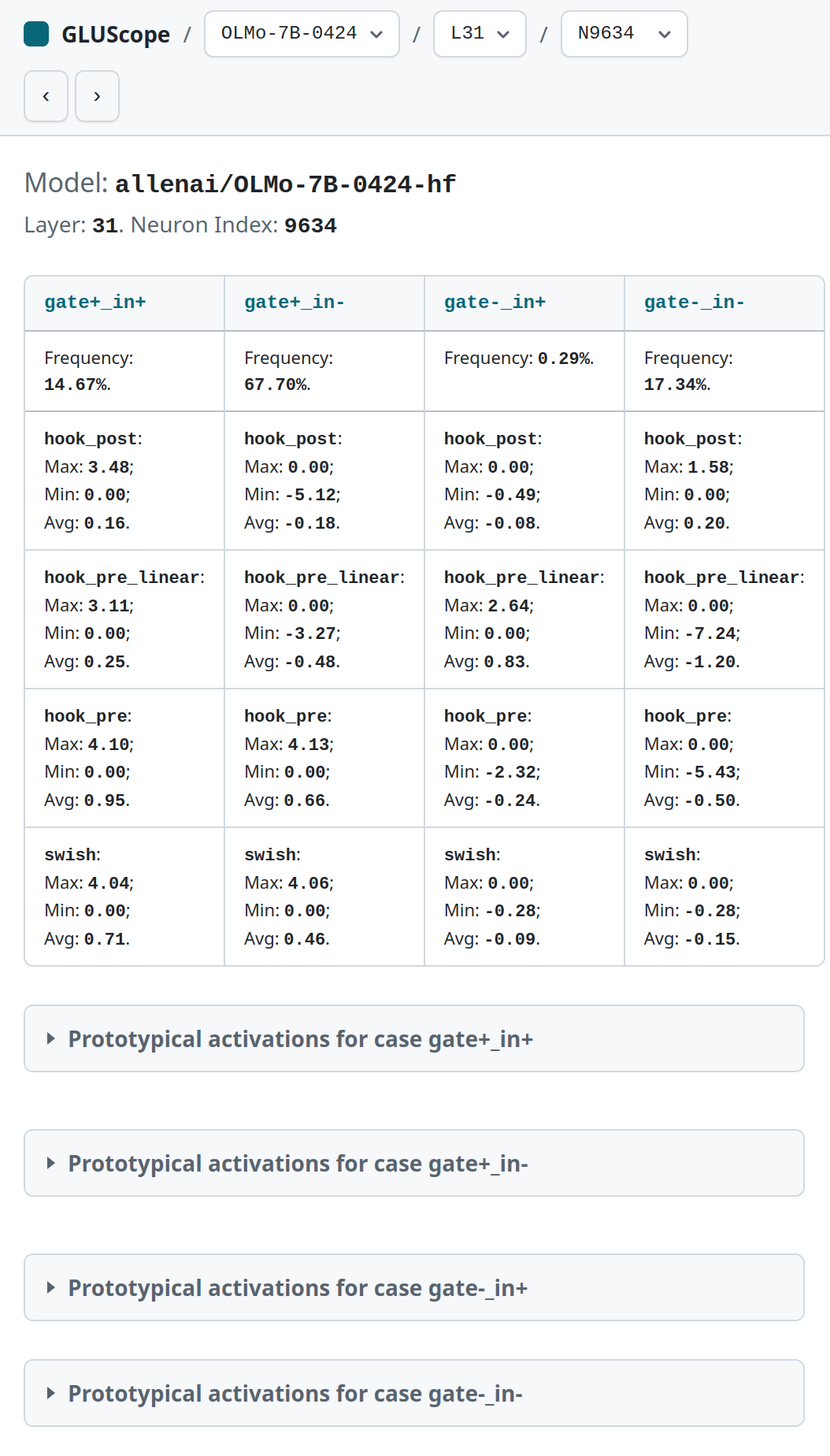}
	\caption{Overview of a neuron page (neuron 31.9634 of OLMo-7B-0424).}
	\label{fig:overview}
\end{figure}

We release the following datasets:
(i) the Dolma subset we ran the model on (\cref{sec:dolma}),\footnote{
	\url{https://huggingface.co/datasets/sjgerstner/dolma-small}
}
and
(ii) a range of datasets (one per model) containing summary information on the activations of each neuron (\cref{sec:activation-data}, see there for the links).
Based on these datasets, we release a website for visualizing neuron data, \neuroscope{}, at
\url{https://gluscope.github.io} (\cref{sec:neuroscope}).
Finally,
the code used to create the artifacts is released at
\url{https://github.com/sjgerstner/gluscope} (\cref{sec:code}).

We release all of the above under the MIT license,
except the Dolma subset which maintains the ODC-BY license of Dolma.

\subsection{Dolma subset}
\label{sec:dolma}
The dataset was produced by the script \texttt{a\_dataset.py} in our code repository.
It is a random subset of Dolma \citep{soldaini-etal-2024-dolma},
a general-purpose LM pretraining dataset.
We made the following design choices:
\begin{itemize}
	\item We keep slightly above 20M tokens.\footnote{
		Token counts are based on the OLMo \citep{groeneveld-etal-2024-olmo} tokenizer, which is essentially the same as GPT-NeoX \citep{Black2022GPTNeoX20B}.
	}
	\citet{Nanda2022neuroscope} used more than this, but \citet{voita-etal-2024-neurons} already found good results with 20M tokens, and we want to keep things comparatively lightweight.
	\item As an additional way to keep things lightweight, we retain at most 1024 tokens per example (instead of 2048 in \citet{Nanda2022neuroscope}). Documents longer than this are truncated.
	\item When tokenizing, we let the first token be EOS
	(as is standard in TransformerLens but not in Huggingface).
	We counted this towards the limits of 1024 tokens per example and 20M+ tokens overall.
\end{itemize}

As a result, the dataset contains 45,736 rows of varying length.
Users can run the same script with different arguments
to get another dataset with different design choices.

By publishing this dataset, we enable other users to efficiently load a small subset of Dolma
(without having to load a bigger dataset and then downsample it).
The primary use case we are thinking of is model analysis research like ours.
As with any text dataset, other use cases can be imagined,
such as pretraining small language models (in the spirit of \citealp{warstadt-etal-2023-findings}).

\subsection{Activation datasets}
\label{sec:activation-data}

These datasets were produced by the script \texttt{b\_activations.py}:
We run a model on a text corpus
and record information about the activations of each neuron.
In the following we describe which models and corpus we use, but
with the same script, users can also compute analogous activation datasets for other models and other corpora.

We used the following pretrained language models (links to the resulting activation datasets are in the footnotes):
OLMo-7B-0424 \citep{groeneveld-etal-2024-olmo},\footnote{
	\url{https://huggingface.co/datasets/sjgerstner/OLMo-7B-0424-hf_neuron-activations}
}
Gemma-2-2b \cite{Gemma2024},\footnote{
	\url{https://huggingface.co/datasets/sjgerstner/gemma-2-2b_neuron-activations}
}
Llama-3.2-3B \cite{Llama3.2},\footnote{
	\url{https://huggingface.co/datasets/sjgerstner/Llama-3.2-3B_neuron-activations}
}
Llama-3.1-8B \cite{Llama3.1}.\footnote{
	\url{https://huggingface.co/datasets/sjgerstner/Llama-3.1-8B_neuron-activations}
}

OLMo-7B-0424 has also been used in some other model analysis works \citep{Gerstner2026Weakeningneurons,hakimi-etal-2025-time}.
We think OLMo models are an especially good choice for this kind of research
because their training dataset is publicly available,
which is not the case for most other open-weights models.

For all of these models we used the same text dataset, the Dolma subset from \cref{sec:dolma}.
This choice is especially appropriate for OLMo, since Dolma precisely represents OLMo's training distribution.
The other models do not have a publicly available training dataset. Some of these models were trained on multilingual datasets, so Dolma (which is English-only) presumably only represents a sub-distribution of their training distribution.

In the following we describe in more detail what kind of activation information the dataset contains.
First of all,
we count how often each activation quadrant (\cref{ss:quadrants}) appears.
Then, within each quadrant, we record detailed information about each of the \textit{intermediate activations} of the neuron
($\xgate, \sigma(\xgate), \xin$,\footnote{
Both Swish and GELU are monotonically increasing on positive inputs, so for the quadrants with $\xgate>0$ we do not separately consider $\sigma(\xgate)$.
}
and the final activation $\sigma(\xgate)\cdot\xin$):
the average value, and
the top-k (or bottom-k, depending on the sign) values together with the dataset indices where they appeared.\footnote{
To be precise: each dataset index appears only once per top-k list, even if the text led to several high activations at different positions.
}
We chose $k=16$, which seemed likely to strike a good balance between showing enough information to be meaningful, and avoiding information overload.\footnote{
For comparison: Neuroscope \citep{Nanda2022neuroscope} shows 20 examples per neuron, but in our case the $k$ is multiplied by 4 activation quadrants and 3 to 4
intermediate activations.
}

This activation dataset has two uses:
First, we use it to produce the neuron visualizations of \cref{sec:neuroscope}.
Second, we can use it as input data for experiments investigating properties of neuron activations:
For example, in \citet{Gerstner2026Weakeningneurons} we used an earlier version
of the OLMo activations dataset to uncover a correlation between frequency
of $\xgate>0$ and a property of neuron weights.
See \cref{sec:dataset-example}.

\subsection{Our demo website: \neuroscope}
\label{sec:neuroscope}
Next to the machine-readable activation dataset of \cref{sec:activation-data},
we want to present neuron information in a human-friendly format.
This is what our \neuroscope{} website does.
It is published at
\url{https://gluscope.github.io}.

The website deliberately shows just a few selected neurons,
since it is impossible for a human to look at every single neuron of a model.
Specifically we show neuron 0 from each layer of each model (which is equivalent to a random selection),
plus all the (OLMo) neurons that were mentioned in \citet{Gerstner2026Weakeningneurons}.
To add a page for another neuron,
a user can run the visualization code on their own.
Thanks to the activation dataset,
this does not require running the model on all of the text again.
Optionally, the user can then open a pull request
on \url{https://github.com/gluscope/gluscope.github.io}
to make the page available to the public.

Each \neuroscope{} page describes a single neuron.
It consists of two main components:
summary statistics
and text examples.
See \cref{fig:overview} for an example neuron page.

\begin{figure}
	\includegraphics[width=\linewidth]{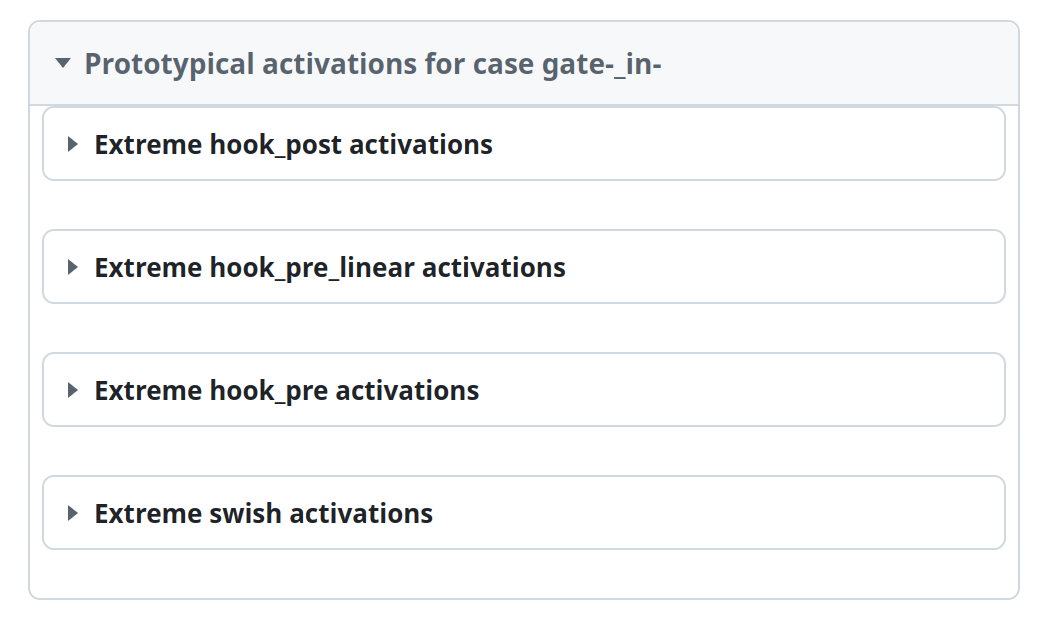}
	\caption{Within each activation quadrant (here, the combination of $\xgate<0$ and $\xin<0$), we show examples for the four intermediate activations \texttt{hook\_post, hook\_pre\_linear, hook\_pre} and \texttt{swish}. See \cref{sec:summary-stats} for definitions.}
	\label{fig:intermediate}
\end{figure}

\subsubsection{Summary statistics}
\label{sec:summary-stats}
The summary statistics are directly read off the activation dataset (\cref{sec:activation-data})
and presented as a table
(as seen in the top half of \cref{fig:overview}).
Each column represents one of the four activation quadrants (\cref{ss:quadrants}).
Within each column, the following information is presented:
\begin{itemize}
	\item Frequency: How often this quadrant appeared when running the model on the dataset.
	\item Detailed information about each of the intermediate activations of the neuron (the terminology here largely follows TransformerLens):
		\texttt{hook\_post} (final activation, i.e. $\sigma(\xgate)\cdot\xin$);
		\texttt{hook\_pre\_linear} ($\xin$);
		\texttt{hook\_pre} ($\xgate$);
		\texttt{swish} ($\sigma(\xgate)$, where $\sigma$ represents either Swish (for most models) or GELU (for Gemma)).
	For each of these, we present the maximum, minimum, and average activations (within the given quadrant).
\end{itemize}

The user can easily infer more coarse-grained statistics.
For example, the frequency of $\xgate>0$ is the sum of the frequencies of \texttt{gate+\_in+} and \texttt{gate+\_in-}.

\subsubsection{Text examples}
\label{sec:texts}
\begin{figure}
	\includegraphics[width=\linewidth]{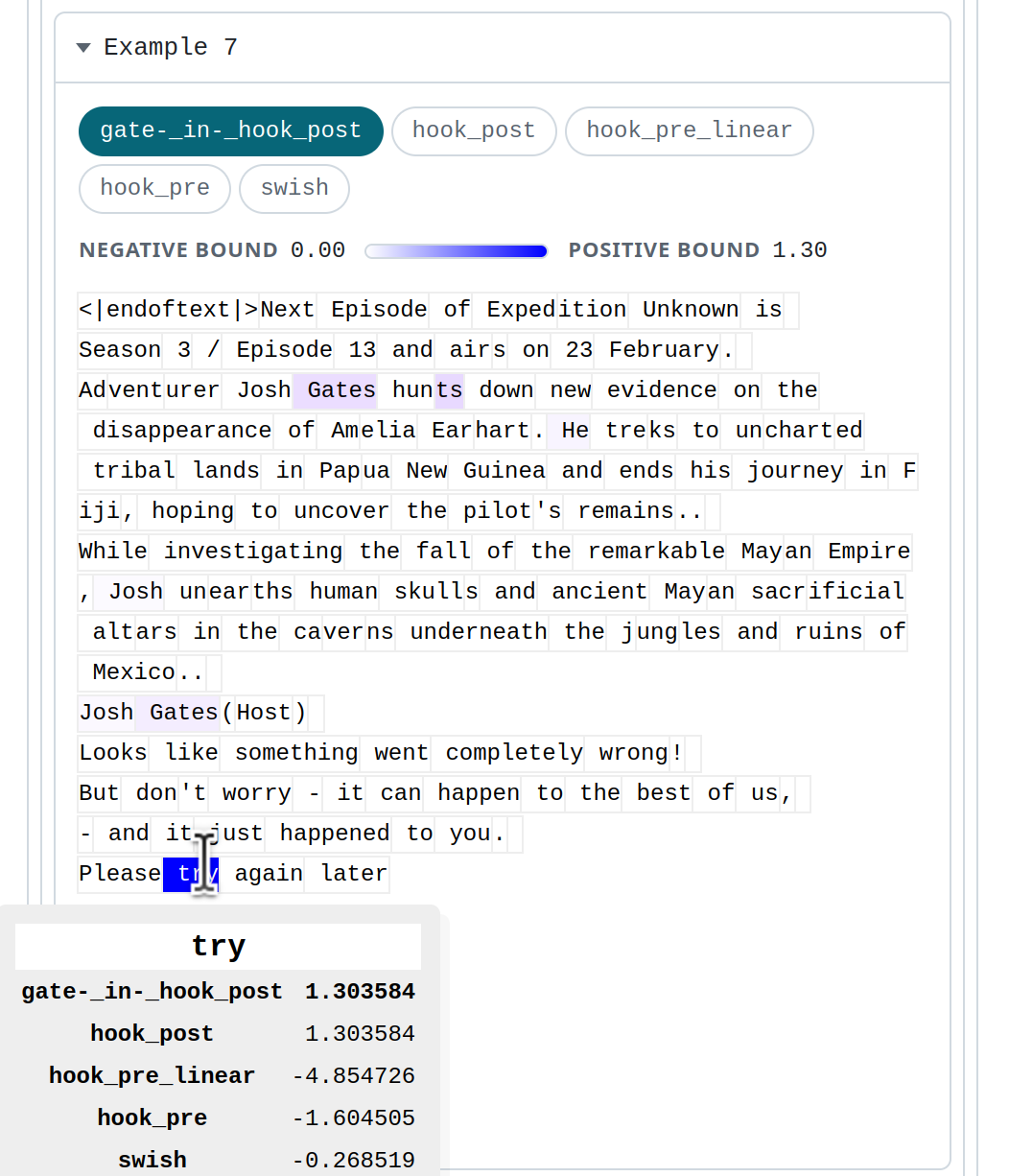}
	\caption{
		A single example of a strong activation of OLMo-7B neuron 31.9634 in the case $\xgate<0$, $\xin<0$.
		The colored tokens are those where this activation quadrant occurred for this neuron.
		Via the top bar it is possible to display all intermediate activations as well.
		Hovering over a token shows the exact activation values.
	}
	\label{fig:single-example}
\end{figure}

The next section is the largest one of each page:
it presents text examples (highest absolute activations) for each of the entries in the summary table.
For example, the first subsection presents those examples that get the largest \texttt{hook\_post} activations, among the examples in the \texttt{gate+\_in+} quadrant.
See \cref{fig:intermediate,fig:single-example}.
Following the activation dataset, we show 16 examples for each entry.

The texts are presented in a truncated version:
We only show them up to the two tokens directly after the token of interest.

\subsection{Code}
\label{sec:code}
We publish the code for generating the datasets and website.
Using this code, researchers can create similar datasets and visualization pages for their own models and corpora.
An example would be to run models on a non-English corpus, to understand if certain neurons behave differently on different languages.
Researchers could also use more specific datasets focusing on a given phenomenon of interest, such as refusal, hallucinations, or biased mechanisms.

\section{Usage examples}
\label{sec:examples}
The goal of the resource is to aid interpretability research.
We present two usage examples
that show how the resource can lead to novel insights.
We already separately described these insights in \citet{Gerstner2026Weakeningneurons}.
Here we focus on how the resource enables these findings,
rather than the findings themselves or their implications.

\begin{table}
	\caption{Pearson correlations between $\cos(\win,\wout)$ and frequency of $\xgate>0$. Sample size is number of MLP neurons in the model. All correlations are significant with $p<0.01$. This table is copied from \citet{Gerstner2026Weakeningneurons}.}
	\label{tab:freq-all}
	\centering
	\begin{tabular}{c|r}
		Model & correlation \\
		\hline
		OLMo-7B-0424 & -0.91 \\
		Gemma-2-2B & -0.76 \\
		Llama-3.1-8B & -0.59 \\
		Llama-3.2-3B & -0.54 \\
	\end{tabular}
\end{table}

\subsection{Using the activation dataset for a model-wide analysis\footnote{
	The insight we find here is described in more detail in section 7 of \citet{Gerstner2026Weakeningneurons}. Here we focus on how \neuroscope{} enables this insight.
}
}
\label{sec:dataset-example}
In this first example,
we demonstrate a use of the activation datasets (\cref{sec:activation-data}).
Specifically, \citet{Gerstner2026Weakeningneurons} find a strong negative correlation between $\cos(\win,\wout)$ and frequency of $\xgate>0$ of any given neuron (see \cref{tab:freq-all}).
The code snippet in \cref{fig:code} shows how to reproduce this finding with our activation datasets.
Users might discover other correlations
between seemingly unrelated properties of neurons
by adapting this code
(e.g. replacing activation frequency with other metrics recorded in our dataset,
and/or replacing $\cos(\win,\wout)$ with other separately computed metrics).

\subsection{Inspecting a neuron with \neuroscope{}\footnote{
		The insights of this section are also described in appendix J.3 of \citet{Gerstner2026Weakeningneurons}. Here we focus on how \neuroscope{} enables these insights.
}}
\label{sec:neuron-example}
Now, in our second example, we demonstrate how the \neuroscope{} website can be used to inspect a given neuron.
We do this for neuron 31.9634\footnote{
We use the notation "layer.neuron", with zero-based indexing. The model has 32 layers, so this neuron is in the final layer.
} of OLMo-7B-0424.

From a weight-based analysis (independent of \neuroscope{})
we know that the neuron's $\wout$ weight closely corresponds to the unembedding representation of "\textit{again}" and a few similar tokens,
while $\wgate$ and $\win$ correspond to "minus \textit{again}".

This naturally leads to the question when the neuron strongly activates:
Is it when "\textit{again}" is a plausible next token (because of $\wout$),
or perhaps when it is \textit{not} a plausible next token (because of $\wgate$ and $\win$)?
Moreover,
we may
find different patterns for different activation quadrants (\cref{ss:quadrants}).

Based on the weights,
since $\wgate$ and $\win$ are highly similar to each other,
we would expect that $\xgate$ and $\xin$ would usually have the same sign,
thus leading to a positive activation most of the time.
However this has to be tested empirically.

\neuroscope{} helps us answer these questions,
leading to the following findings:

\shortpar{Frequent negative activations}
Contrary to our prediction based on the weights,
the neuron often activates negatively.
Specifically,
$67.7\%$ of all activations are of the type $\xgate>0, \xin<0$.
See the table in \cref{fig:overview}.

\shortpar{Most activation quadrants are hard to interpret}
For the \texttt{gate+\_in+} and \texttt{gate-\_in+} quadrants,
we do not find any obvious pattern in the activations.
For example, the strongest \texttt{gate+\_in+} activation is on the token "\textit{door}" (next token "\textit{often}"),
but the next-strongest one is on the "\textit{es}" of "\textit{volcanoes}" (next token "\textit{body}").

For the \texttt{gate+\_in-} activations we do find a pattern,
but it is not obviously related to "\textit{again}":
The next token is usually an adverb like "\textit{meanwhile}", "\textit{instead}", "\textit{later}".

\shortpar{The \texttt{gate-\_in-} activations are highly interpretable}
When we look at this fourth activation quadrant
(which only appears $17.34\%$ of the time),
we suddenly find that the largest of these activations are closely related to "\textit{again}":
Often in these cases "\textit{again}" is the correct next token,
or at least a plausible one.
For example, these activations often occur on the token "\textit{once}",
as in "\textit{once again}".
See \cref{fig:single-example} for another example.
In these cases, ``again'' is already highly ranked,
but a different token may be generated by the model.
This neuron increases the strength of the ``again'' direction and
makes it more likely that the correct token ``again'' is generated.

We could not have discovered this last point with
a more traditional resource,
like those of \citet{Nanda2022neuroscope} or \citet{Choi2024Scalingautomaticneuron}:
If we had just recorded the largest positive activations
(and possibly the largest negative ones)
we would only have got those from the case \texttt{gate+\_in+} (and \texttt{gate+\_in-}).
(As shown by the \neuroscope{} page in \cref{fig:overview},
their extrema are at $3.48$ and $-5.12$.)
The highly interpretable \texttt{gate-\_in-} activations
are much smaller (at most $1.58$)
and would therefore have been missed.

\section{Conclusion}
We release \neuroscope{},
a resource for inspecting neurons in transformer language models.
We fill an important gap of previous resources
by focusing on GLU-based activation functions
and giving equal consideration to all of their four activation quadrants.
Our resource includes (i) datasets of neuron activations spanning all four quadrants; (ii) a website for visualizing this data; and (iii) code for creating similar datasets and visualizations.
Thanks to the new focus on GLU and activation quadrants,
our resource enables more fine-grained analyses than previous ones,
and surfaces interpretable neuron behavior that other tools would miss.

\section*{Limitations}
Focusing on gated activation functions addresses an important limitation of previous work,
but is itself limited:
Our tool cannot directly be used to analyze
non-Transformer models like Mamba \citep{Gu2024Mamba}.
We also did not test our code for larger models requiring several GPUs, or for mixture-of-experts (MoE) models. Our code would require adaptations for these use cases.

Moreover, we focus on individual neurons,
instead of other hypothesized causal variables like SAE features \cite{Sharkey2022}.

A limitation specific to the analysis of non-OLMo models is that our Dolma subset does not match their training distribution, and only represents a sub-distribution of it. This may lead to interpretability illusions \cite{Bolukbasi2021interpretabilityillusionBERT}.
In particular Dolma (and hence OLMo) only contains English and code, whereas the Llama models were trained on multilingual datasets. Gemma 2 is also primarily English and code, so here it is less of a limitation.

We mainly care about showing all four activation quadrants of a neuron,
and focus less on other aspects.
In particular, contrary to e.g. \citet{Choi2024Scalingautomaticneuron}, we do not include LLM-generated explanations.

\bibliography{custom,anthology}

\appendix

\section{Author contributions}
The idea for this paper was developed together by S.G. and H.S.

S.G. did the most of the programming and paper-writing.
L.V. improved the user experience of the website, and provided valuable feedback on the writing of the paper.
H.S. supervised the project.

\section{Acknowledgments}
We thank Lea Hirlimann for valuable feedback and encouragements.

This work was funded by Deutsche Forschungsgemeinschaft (project SCHU 2246/14-1).

The authors gratefully acknowledge the scientific support and HPC resources provided by the Erlangen National High Performance Computing Center (NHR@FAU) of the Friedrich-Alexander-Universität Erlangen-Nürnberg (FAU) under the NHR project b309dd / JA-27440 / InterpGLU. NHR funding is provided by federal and Bavarian state authorities.

\section{Deontology statements}\label{ap:responsible}
\subsection{Licenses and languages of models and data}

\shortpar{Dolma and OLMo}
Training and inference code, weights (OLMo), and data (Dolma) under Apache 2.0 license.
``The Science of Language Models'' is explicitly mentioned as an intended use case.
Dolma is quality-filtered and designed to contain only English and programming languages
(though we came across some French sentences as well)
\cite{groeneveld-etal-2024-olmo,soldaini-etal-2024-dolma}.

\shortpar{Gemma}
To download the model one needs to explicitly accept the
\href{https://ai.google.dev/gemma/terms}{terms of use}.
NLP research is explicitly listed as an intended usage \citep{Gemma2024}.

Gemma 1 is primarily English and code \citep{Gemma2024}; Gemma 3 was also trained on multilingual data \citep{Gemma3}. We could not find any information on the training data of Gemma 2.

\shortpar{Llama}
Inference code and weights under an ad hoc
\href{https://github.com/meta-llama/llama/blob/main/LICENSE}{license}.
There is also an
\href{https://github.com/meta-llama/llama/blob/main/USE_POLICY.md}{``Acceptable Use Policy''}.
Our work is well within those terms.

In Llama 1,
languages mostly include English and programming languages,
but also Wikipedia dumps from
``bg, ca, cs, da, de, en, es, fr, hr, hu, it,
nl, pl, pt, ro, ru, sl, sr, sv, uk''
\citep{Touvron2023}.
In Llama 3.1 and 3.2, supported languages are English, German, French, Italian, Portuguese, Hindi, Spanish, and Thai \citep{Llama3.1,Llama3.2},
but at least Llama 3.2 has been trained on a broader range of languages without officially supporting them \citep{Llama3.2}.
There is no more detailed public information about the training data.

\subsection{Computational complexity}

The GPU used was a single NVIDIA A100 80GB.
In the following, time units refer to this GPU type.

\shortpar{Activation datasets}
Generating an activation dataset from around 10 hours for Gemma-2-2B to around 20 hours for Llama-3.1-8B.s

For OLmO-7B, we used an earlier, much less efficient version of the code, and the run took approximately five days. We expect that with the current version it would take at most 24 hours.

\shortpar{Neuron pages}
After the activation dataset is computed,
generating a single neuron page takes up to 1-2 minutes, depending on the layer (layer 0 pages just take a few seconds).

\subsection{LLM use}
We used LLMs for the following tasks:
\begin{itemize}
	\item Help with programming and web design. The JavaScript code of the website is LLM-generated and was not carefully checked.
	Python code is mostly our own work, and any LLM-generated snippets were manually checked.
	\item Ideas for ways to release the system.
	\item Some formulations for the paper. The paper is still mostly human-written.
\end{itemize}

The ideas presented in this paper are our own.

\section{Additional figures}
\Cref{fig:code}.

\begin{figure*}
	%
	%
	%
	%
	%
	
	\begin{lstlisting}[language=Python,breaklines]
		#load precomputed cos(w_in,w_out) data,
		#assumed format is a tensor of shape (layer, neuron).
		cosines = torch.load('cosine_data.pt')
		
		#load our activation dataset
		activation_dataset = datasets.load_dataset('sjgerstner/OLMo-7B-0424-hf_neuron-activations')['train']
		#compute frequency of x_gate>0 for each neuron
		freq_gate_pos = torch.Tensor(activation_dataset['gate+_in+_freq']) + torch.Tensor(activation_dataset['gate+_in-_freq'])
		
		#compute correlation and p value
		corr_and_p = stats.pearsonr(cosines, freq_gate_pos)
	\end{lstlisting}
	
	\caption{Essential coding steps for \cref{sec:dataset-example}}
	\label{fig:code}
\end{figure*}

\end{document}